\documentclass[twoside]{article}
\usepackage{amssymb}
\usepackage{graphicx}
\usepackage{times}
\usepackage{fuzz}

\def\typeproof{}                                                

\begin{document}

\title{LSIS Research Report 2003-006\\
Modeling Object Oriented Constraint Programs in Z}

\def\classification{Artificial Intelligence, Constraint Programming, Formal Specification}
\def\keywords{Artificial Intelligence, Constraint Programming, Formal Specification, Z, UML}

\author{Laurent Henocque\\
Laboratoire des Sciences de l'Information et des Syst\`emes\\
LSIS (UMR CNRS 6168)\\
Campus Scientifique de Saint J\'er\^ome\\
Avenue Escadrille Normandie Niemen\\
13397 MARSEILLE Cedex 20}

\maketitle

\begin{abstract}
Object oriented constraint programs (OOCPs) emerge as a leading evolution of constraint programming and artificial intelligence, first applied to a range of industrial applications called configuration problems. The rich variety of technical approaches to solving configuration problems (CLP(FD), CC(FD), DCSP, Terminological systems, constraint programs with set variables, \ldots) is a source of difficulty. No universally accepted formal language exists for communicating about OOCPs, which makes the comparison of systems difficult. We present here a Z based specification of OOCPs which avoids the falltrap of hidden object semantics. The object system is part of the specification, and captures all of the most advanced notions from the object oriented modeling standard UML. The paper illustrates these issues and the conciseness and precision of Z by the specification of a working OOCP that solves an historical AI problem : parsing a context free grammar. Being written in Z, an OOCP specification also supports formal proofs. The whole builds the foundation of an adaptative and evolving framework for communicating about constrained object models and programs.
\end{abstract}

\newtheorem{theorem}{\sffamily Theorem}
\newtheorem{proposition}{\sffamily Proposition}
\newtheorem{lemma}{\sffamily Lemma}
\newtheorem{corollary}{\sffamily Corollary}
\newtheorem{example}{\sffamily Example}
\newtheorem{remark}{\sffamily Remark} 
\newtheorem{definition}{\sffamily Definition}
\newtheorem{conjecture}{\sffamily Conjecture}
\def\proof{{\sc Proof.}\quad }

\def\qed{\quad{$\blacksquare$}\medskip }
\def\qedo{\quad{$\square$}\medskip }

\label{firstpage}
\section*{Introduction}
\subsection*{From Configuration to Object Oriented Constraint Programs}
Rule based systems, logic programming, and a recent evolution of constraint programming have been applied to a category of problems called \emph{configuration} problems. 
Configuring means simulating the construction of a composite and complex product, based on a library of elementary components. Components are subject to relations (this is the \emph{partonomic} information), and participate to inheritance relationships (this is called the \emph{taxonomic} information). 
Given an input in the form of a partial product and specific constraints, the goal of configuration is to pick up or generate, then interconnect the necessary components, for finally deciding upon their exact type and attribute values. Configuration output is a complex interconnected product respecting well formedness rules stated by various constraints. This combinatorial problem is explicitly formulated as a finite model generation problem.

\begin{figure}[htb]
   \centering
      \includegraphics[angle=270,width=\textwidth]{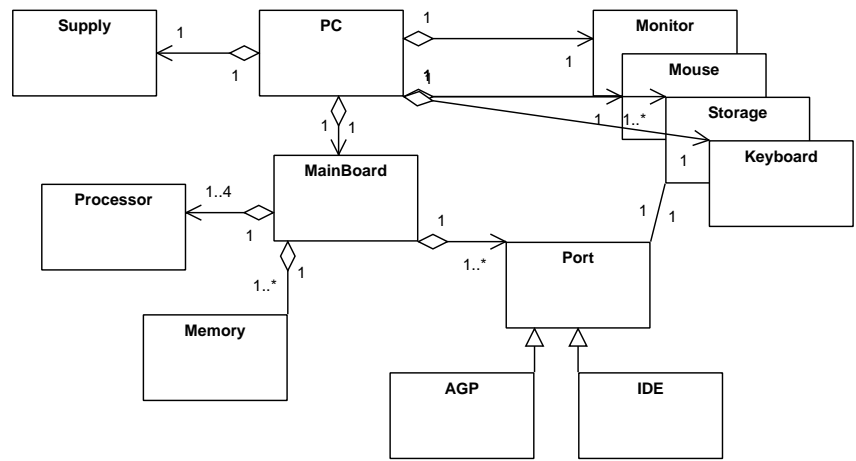}
    \caption{A simplified object model for a personal computer}
    \label{fig:computer}
\end{figure}

The figure \ref{fig:computer} illustrates this with a classical example. Configuring a personal computer (PC) consists in picking components from a catalog of component parts (e.g. processors, hard disks in a PC ), using known relations between types (motherboards can connect up to four processors), and instantiating object attributes (selecting the ram size, bus speed, \ldots). 
Constraints apply to such configuration problems that define which products are valid, or well formed. For example in a PC, the processors on a motherboard all have the same type, the ram units have the same wait times, the total power of a power supply must exceed the total power demand of all the devices. 
Configuration applications deal with such constraints, that bind variables occurring in the form of variable object attributes deep within the object structure.

We suggest to abandon the term configuration, bound to a very specific application area (even if it is broadly distributed in the economy), in favor of a more general purpose and AI related denomination : object oriented constraint programs (OOCP for short). OOCP has many potential AI applications, ranging from context free language parsing (we use this example), to image recognition, or distributed agent intelligence and planning.  

\subsection*{Existing approaches}
The industrial need for configuration  is widespread, and has triggered the development of many 
applications, as well as generic configuration tools or configurators, built upon all available technologies. For instance,
configuration is a leading application field for rule based expert systems. As an evolution of R1\cite{McDermott82}, the
XCON system \cite{barkerconnor89} designed in 1989 for computer configuration at Digital Equipment involved  31000
components, and 17000 rules. The application of configuration is experimented or planned in many different industrial
fields, electronic commerce (the CAWICOMS project\cite{Felfernig02}), software\cite{Ylinen:02}, computers\cite{Plain2002},
electric engine power supplies\cite{JohnGeske99} and many others like vehicles, electronic devices, customer relation
management (CRM) or even software\cite{Ylinen:02,FSB99}.

The high variability rate of configuration knowledge (parts catalogs may vary by up to a third each year) makes
configuration application maintenance a challenging task. Rule based systems like R1 or XCON lack modularity in that
respect, which encouraged researchers to use variants of the CSP formalism (like DCSP
\cite{Mittal:90AA,soininen99fixpoint,AFM02}, structural CSP \cite{Nareyek:99}, composite CSP
\cite{sabinfreudercomposite96}), constraint logic programming (CLP \cite{JL87}, CC \cite{FSB99}, stable model semantics for logic programs with disjunctions
\cite{soininen01representing}), or object oriented approaches\cite{Stumptner97,Stumptner98AIEDAM,Mailharro:98,Meyer99}. 

Because of the variety of approaches to this problem (CLP(FD), CC(FD), CP and extensions, Expert Systems), no common language is available for researchers to exchange problem statements and compare their results. Each of the above cited articles uses an ad-hoc description of its working example. Some UML\cite{UML} models are presented from time to time, which never allow to overcome the ambiguities inherent to this exercise, even though the UML is far more formal than people usually think. The objective of this paper is to propose to this growing community a common language for exchanging models and problems.

\subsection*{Paper objectives}

This paper presents a general use object oriented constraint system for the specification of object oriented constraint problems. The object system is not predefined, or provided as an object oriented extension to some specification language. Instead, we have chosen to make the object system specification explicit, using the Z language \cite{zspivey}. There are several reasons for both the choice of Z and this approach: 
\begin{itemize}
\item we feel that in order to be widely accepted, the underlying semantic of an object system must be questionable, commented, improved, and formally established, 
\item the Z language has very simple and clear semantics, and offers an extremely rich range of relational operations, a crucial issue in object oriented constraint programming,
\item the Z language was shown to have the favor of the industry over other specification language\cite{SS-ZUM91}, essentially because of its structure (the grouping concept introduced by schemas)
\item we have succeeded in specifying in Z the most advanced class modeling constructs from the UML\cite{UML}, which has become the standard in object oriented modeling. This guarantees that modeling cannot be biased, or tweaked by arbitrary limitations in the object language, and that any object oriented model can be specified
\item Z specifications can be type checked (we used \fuzz\footnote{available at http://spivey.oriel.ox.ac.uk/~mike/fuzz/}\ extensively to type check this paper), which offers a first level of formal verification, 
\item Z specifications are subject to formal proofs, possibly assisted or automatized by theorem provers,
\item Z offers built in extensibility features, that allow to formally define, then use, any operator or relation using any possible syntax (prefix, infix, postfix). We exploit this feature to improve the readability of constraints.
\end{itemize}
To summarize this, we found that Z is the simplest logic offering both structure (via schemas) and support for the formal definition of complex structural constraints or expressions. This last issue is crucial to object oriented constraint programs, for instance to allow the statement of a constraint relating, in a personal computer, the sum of powers used by all elementary electrical devices, and the power made available by the supply. It was furthermore argued\cite{jacobs99coalgebras} that coalgebras support most of the notions required to deal with object state and class invariants. Relational algebraic languages like Z provide the capacity of specifying both algebras (types and their operations) and coalgebras (states and their transitions).

There have been many attempts to capture object orientation within specification languages, either viewed as Z extensions (Object Z\cite{objectz}, OOZE\cite{ooze91}) or based upon other mathematical grounds (the FOOPS\cite{borba94operational} extension to OBJ). A logical approach to object orientation is also put to work in terminological knowledge representation languages\cite{klone85,baader91terminological}). Also, constraint programming has been introduced in object oriented knowledge representation languages (as e.g in CLAIRE\cite{caseau94constraint}). 

We are not presenting the latest object oriented language, or system, or extension to whichever existing approach. Our aim is to capture rich enough object oriented semantics in a simple and unmodified logic (hence Z), rather than to rely upon the inherent semantics of an object oriented extension of some logic. By doing so, we prevent both the potential expressiveness limitations of any given object system, and the possibility that its semantics are improperly defined, or questionable. Our approach allows to document an object oriented constraint program, by simultaneously specifying both the object system semantics, and the problem itself. 
This task is made simpler because we do not need to specify state transitions (coalgebra operations), but reason exclusively about state. Essential issues in object oriented \emph{programming} like polymorphism, or concurrency are irrelevant here. Our goal is to express valid object system states, using constraints, which altogether describe an object oriented constraint program. This simplifies the use we make of Z, because in our case decorations are useless.

The paper is organized as follows : section 1 briefly introduces essential aspects of Z. Section 2 specifies the class and type features of an object system,   illustrating how all essential object oriented modeling concepts can be captured. Section 3 specifies relations and roles. Section 4 details how object structure constraints can be formulated, and introduces useful auxiliary constructs. Section 5 presents the specification of an artificial intelligence application of object oriented constraint programs to context free grammars parsing. Section 6 is the conclusion. It can be a possible reading strategy to first take a glance at section 5, since it illustrates the essential motivation of this work.

\section{Introducing Z}

For space reasons, it is impossible to make this paper self contained, since this would suppose a thorough presentation of both the UML notation\cite{UML}, and the Z specification language\cite{zspivey}. The reader, if novice in these domains, is kindly expected to make his way through the documentation, which is electronically available. 
For clarity however, we provide a brief description of several useful Z constructs. More advanced notations or concepts will be introduced when necessary. 
\subsection{data types as named sets}
Z data types are possibly infinite sets, either uninterpreted,  defined axiomatically, or defined as (possibly recursive) free types as the next three examples illustrate. 
\begin{zed}
[DATE]
\end{zed}
\vspace{-0,7cm}
\begin{axdef}
dom:\finset \nat
\end{axdef}
\vspace{-0,7cm}
\begin{zed}
	colors ::= red | green | blue
\end{zed}
\noindent From this on, all possible relation types can be built from cross products of other sets. 

\subsection{axiomatic definitions}
Axiomatic definitions allow to define global symbols having plain or relation types. For instance, a finite group is declared as 

\begin{axdef}
zero:dom\\
inverse:dom \fun dom\\
sum:dom \cross dom \fun dom\\
\where
\forall x:dom @ sum(x,inverse(x))=zero\\
\forall x:dom @ sum(x,zero)=x\\
\forall x,y:dom @ sum(x,y)=sum(y,x)\\
\forall x,y,z:dom @ sum(x,sum(y,z))=sum(sum(x,y),z)\\
\end{axdef}

\noindent The previous axiomatic definition illustrates cross products and function definitions as means of typing Z elements. Now axioms or theorems are expressed in classical math style, involving previously defined sets. For instance, we may formulate that the inverse function above is bijective (this is a theorem) in several equivalent ways as e.g.:
\begin{zed}
inverse \in dom \bij dom\\
\forall y:dom@\exists_1 x:dom @ inverse (x)=y
\end{zed}
\subsection{schemas}

\noindent The most important Z construct, \emph{schemas}, occur in the specification in the form of named axiomatic definitions. A schema $[D|P]$ combines one or several variable declarations (in the declaration part $D$) together with a predicate $P$ stating validity conditions (or constraints) that apply to the declared variables.

\begin{schema}{SchemaOne}
a : \nat\\
b : 1 \upto 10
\where
b < a
\end{schema}

\noindent The schema name hides the inner declarations, which are not global. A schema name (as $SchemaOne$ above) is used as a shortcut for its variable and predicate declarations that can be universally or existentially quantified at will. Schemas are $true$ or $false$ under a given \emph{binding}. For instance, $SchemaOne$ is $true$ under the binding $\langle 4 \bind a, 3 \bind b \rangle$ and $false$ under the bindings $\langle 3 \bind a, 234 \bind b \rangle$ or $\langle 3 \bind a, 4 \bind b \rangle$. The latter violates the explicit constraint stated in the predicate part of the schema, while the former also violates the implicit constraint carried by the interval definition $1 \upto 10$ (a subset of $\nat$). In some contexts, a schema name denotes the set of bindings under which it is true.

Z allows Boolean schema composition. Two schemas can be logically combined (e.g. "anded") by merging their declaration parts provided no conflict arises between the types of similarly called variables, and by applying the corresponding logical operator (e.g. the conjunction) to the predicates. For instance, given the schema $SchemaOne$ above, and another schema called $SchemaTwo \defs [b:\nat; c:\nat | b < c]$ (this illustrates another syntax for simple schema declarations), 
we may form the schema $SchemaThree$, as
\begin{zed}
	SchemaThree \defs SchemaOne \land SchemaTwo
\end{zed}
Incidentally, the variable declarations $b$ in both schemas collide, but not for their types since $b$ is a member of $\nat$ in both cases. The first declaration of $b$ bears a built in constraint, which can be moved to the predicate part. Hence the schema $SchemaThree$ would list as :
\begin{schema}{SchemaThree}
a,b,c : \nat\\
\where
1<b<10\\
b<a\\
b<c
\end{schema}

\section{Classes}

We wish to describe Z specified object oriented constructs so as to reach an expressive power comparable to that allowed by the UML\cite{UML} class (and state) diagrams, hence allowing to model in a purely formal way the static properties of an object constraint system. In order to sit our definitions on clean formal grounds, we propose a generic Z specification that captures all required concepts. In defining classes, we specially need to cover two essential notions : multiple inheritance and multiple discriminator specialization. While the former is attained by existing object oriented Z extensions, the latter is not, which partly justifies this work. An essential contribution of this work is that the object system specification is explicit, and can be discussed, adapted, or extended at will. Essential in this respect is the clarity and soundness of the notion of "object references", made explicit here.

The \emph{schema} notation can be understood as an heterogeneous aggregate of mathematical variables, subject to built in constraints. In other words, schemas can be seen as mathematical variables representing all the possible states of Pascal records, or of C structs. 
From an object oriented point of view, the predicate part of a schema forms the essence of what is called in OO programming a class invariant : a property that must be true of object instances at all times (i.e. before and after any method call).

\subsection{preliminary definitions}
The object oriented vocabulary involves common and rather vague words. We wish to make things precise, and to avoid difficulties in the sequel. A \emph{class definition} holds the description of \emph{class specific attributes} and \emph{class specific invariant} together with inheritance relationships. Accounting for inheritance yields a description of the \emph{(full) class structure} and \emph{(full) class invariant}  which together form the \emph{class specification}. A class \emph{instance}, or \emph{object} is a binding of values to all attributes in the class structure that satisfies the class invariant. 
Such a binding is sometimes referred to as \emph{state}. 
The set of all object instances bijectively maps to a set of object \emph{references}. The bijection between object references and class instances allows for a precise definition of \emph{class} and \emph{type}. We call \emph{class} the set of object references mapped to all the instances of a given class structure. A \emph{type} recursively defines as the union of a given class, and of all its subclass types down the inheritance directed acyclic graph. By defining types as sets, we stay respectful of Z's terminology, which identifies sets and types. All these definitions will be illustrated and made understandable by examples in the sequel.
Z provides enough constructs to account for classification mechanisms. We first define a set $ObjectReference$ of object references as an uninterpreted data type. 

\begin{zed} [ ObjectReference ] \end{zed}
$ObjectReference$ would be interpreted on the set of natural numbers (or a finite subset) by an automated theorem proving approach based on finite model generation. Practical implementations of object systems typically use pointers or integers as object references.
We define $ReferenceSet$ as an abbreviation for finite sets of object references, later used to model object \emph{types}.

\begin{zed}
    ReferenceSet == \finset ObjectReference
\end{zed}

\noindent We define class names as global names using Z's free type declaration syntax. For practical reasons, if a class should have the name $Engine$, we reserve the symbol $Engine$ to denote the corresponding type. The global symbol denoting the class definition is obtained by prepending the string "Class" to the actual name. In our example, the class name thus writes as $ClassEngine$. Depending upon the context, the declaration of class names may look like :

\begin{zed}
	CLASSNAME ::= ClassPC | ClassPrinter | ClassMonitor | \dots
\end{zed}
\vspace{-1,1cm}
\begin{zed}
	CLASSNAME ::= ClassCar | ClassWheel | ClassEngine | \dots
\end{zed}


\noindent We now declare $ObjectDef$ as a predefined super class for all future classes. Object definitions will be used to bijectively map each object to a unique individual from the set $ObjectReference$. Object references are needed in addition to object state since in object oriented modeling two distinct objects may share the same attribute values (whereas in Z  two "bitwise equal" bindings represent the same logical entity). Also, since two distinct Z schemas may have the same Z type, we need to integrate the actual class name in objects.

\begin{schema}{ObjectDef}
i:ObjectReference\\
class:CLASSNAME\\
\end{schema}

\noindent We define a function $instances$ mapping class names to the set of instances of that class.

\begin{axdef}
instances:CLASSNAME \fun ReferenceSet
\end{axdef}

\subsection{defining class structures using inheritance}

\noindent An essential aspect of object oriented modeling is that objects are associated with \emph{state}. On the one hand, inheritance relations allow to restrict the possible values of attributes declared in superclasses (this phenomenon is called inheritance for \emph{specialization}). On the other hand, classes may extend the list of attributes defined in superclasses by their own (this is called inheritance for extension). Most situations where inheritance occurs combine both cases in a single inheritance relation. Z offers built in representation of state in the form of schemas. We now show a way to associate such schemas to individual types in a standardized way, so as to bind state to the types as declared previously. We illustrate this through a simple three class example : class $B$ inherits $A$, extending it with an extra attribute, and class $C$ specializes $A$ with an extra constraint.

Each class $X$ is implemented via two constructs. First, the class definition occurs as a schema called $ClassDefX$ (we prepend "ClassDef" to the desired class name to form the schema name). This schema defines both the class attributes and inheritance relationships, as would any class definition do in object oriented modeling or programming languages. The predicate part of the schema offers room for the specification of class invariants.

\begin{schema}{ClassDefA}
a:1 \upto 10
\end{schema}
\vspace{-0,7cm}

\begin{schema}{ClassDefB}
ClassDefA\\
b:\nat_1
\end{schema}
\vspace{-0,7cm}


\begin{schema}{ClassDefC}
ClassDefA \\
\where
a \geq 5\\
\end{schema}

\noindent Class definitions as seen above account for inheritance by simply copying the definition schemas of the inherited (super) classes. Doing so allows to state constraints involving attributes pertaining to super classes (this is \emph{specialization}). All the predicates present in the inherited classes are conjoined (i.e. logically "anded") to the predicate part of the resulting schema. This formulation hence adequately accounts for both types of inheritance : extension and specialization. Note that the types of all inherited attributes must match. If attributes having the same name are inherited from two distinct superclasses, either or both can be renamed to prevent clashes, as e.g. in :

\begin{schema}{ClassDefD}
ClassDefB[d/b]\\
\where
d=5
\end{schema}
where the constraint $d=5$ actually binds the attribute originally declared as $b$ in $ClassDefB$. A Z type checker can detect errors in the formulation of such class definitions, specially when type conflicts occur for attributes having the same name. A potential cause of error remains if two conceptually distinct yet non distinguishable attributes exist in two inherited classes.

To implement a working object system, we need to add some extra technical information to class definition schemas : object and class references. Like before, schema composition with the logical operator $\land$ offers the expected semantics of extension by combining the schema types of the two schemas and of specialization by conjoining their predicate parts. Assuming the same toy example as before, ($A$ is a toplevel class that $B$ and $C$ inherit), we write the following :

\begin{zed}
	ClassSpecA \defs ClassDefA \land [ObjectDef | class = ClassA~]\\
	ClassSpecB \defs ClassDefB \land [ObjectDef | class = ClassB~]\\
	ClassSpecC \defs ClassDefC \land [ObjectDef | class = ClassC~]\\
\end{zed}

\noindent It must be understood that the schema types corresponding to $ClassSpecA$ and $ClassSpecC$ are the same (this schema type is noted $\lblot i:ObjectReference;a:\nat;class:CLASSNAME \rblot$ in Z), even though the schema names differ, because class $C$ specializes $A$ but does not extend it. Hence a specific workaround is needed to make sure that the set of bindings that satisfy $ClassSpecC$ is not included in $ClassSpecA$, and more generally that no two sets of bindings satisfying two distinct class definition schemas can intersect. This goal is achieved thanks to the $class$ attribute inserted via the schema $ClassDef$, that takes a distinct value for each class.


\subsection{defining class types}

We can now make our toy $ABC$ model more complete, and define what the types $A,\ B,\ C$ represent. We use an axiomatic definition of three sets $A,\ B,\ C$ as finite sets of object references:

\begin{axdef}
A,B,C:ReferenceSet\\
\where
A = instances(ClassA) \cup B \cup C\also
instances(ClassA)=\{o : ClassSpecA | o.class=ClassA @ o.i \}\\
instances(ClassB)=\{o : ClassSpecB | o.class=ClassB @ o.i \}\\
instances(ClassC)=\{o : ClassSpecC | o.class=ClassC @ o.i \}\also
\forall i:instances(ClassA) @  (\exists_1 x : ClassSpecA @ x.i=i)\\
\forall i:instances(ClassB) @  (\exists_1 x : ClassSpecB @ x.i=i)\\
\forall i:instances(ClassC) @  (\exists_1 x : ClassSpecC @ x.i=i)\\
\end{axdef}

\noindent The declaration part in this axiomatic definition declares the type sets corresponding to all the classes in our toy model. The properties of these sets are stated by several axioms : 
\begin{itemize}
\item The types of sub classes are subsets of a class type. A type is the union of (the object references of) all the corresponding class instances, and of the types of its subclasses. Note that by making our example more complete the types $B$ and $C$ might intersect, because of multiple inheritance.
\item The set $instances(ClassX)$ holds the schema bindings having the "class" attribute set to $ClassX$. These sets are pairwise disjoint by construction
\item The other set, $X$, corresponds to the classical notion of a type. 
\item The same object reference cannot be used for two distinct objects in  the same class.
\end{itemize} 
These axioms ensure that each object reference is used at most once for an object. Alternatively stated, no two distinct "object" bindings share the same object reference.
The preceding type definitions make the set $ObjectReference$ the most general type in the model.
Based upon these definitions, any class located in the middle of the inheritance tree is \emph{concrete} : in an interpretation, we may have an instance of $A$ that is neither an instance of $B$ nor $C$.
Finally, this specification makes clear the distinction between :
\begin{itemize}
\item  a class \emph{definition} : this is the schema $ClassDefX$, which accounts for inheritance,
\item a class \emph{specification} : this is $ClassSpecX$, which accounts for object and class references,
\item a \emph{class} : this is represented by the set $instances(ClassX)$,
\item an object's \emph{type} : any set $X$ to which the object's reference belongs (if an object is an instance of class B, and B inherits A, it is accepted to say that this object is a "B", and also that it is an "A").
\end{itemize} 

\subsection{semantics, interpretations, objects}
An \emph{interpretation} of a Z specification is a set of bindings with types corresponding to the schema types that occur in the specification. A \emph{model} in the logical sense is an interpretation that satisfies all the axioms.   
An \emph{object} is a binding satisfying the schema type and properties of some class specification schema ($ClassSpecX$). Note that such a binding may satisfy the schema types and properties of several distinct class specification schemas, because in Z the schema name isn't part of the schema type). This does no harm however, since the $class$ attribute in class specification schemas sorts things apart.
Note that the final axioms in the axiomatic definition of types ($\forall i:instances(ClassX) @  (\exists_1 x : ClassSpecX @ x.i=i)) \dots$ ) constrain the valid interpretations so that each object reference occurs only once among the whole set of object bindings, which to the best of our knowledge cannot be formulated more concisely.

\subsection{creating objects}
Although we do not focus here on the dynamics of the object systems, but rather on the mathematical properties of their valid states, it is of some interest at this point to mention that we are modeling a system that could be specified further to model a practically usable application, where object instances can be created, and destroyed. To achieve this requires to get a hold over the global system state. This is achieved by placing the type definitions within a schema, instead of keeping them as global axiomatic definitions :

\begin{schema}{ObjectSystemABC}
A,B,C:ReferenceSet\\
ObjectsA:\power ClassSpecA
\where
A = instances(ClassA) \cup B \cup C\also
instances(ClassA)=\{o : ClassSpecA | o.class=ClassA @ o.i \}\\
instances(ClassB)=\{o : ClassSpecB | o.class=ClassB @ o.i \}\\
instances(ClassC)=\{o : ClassSpecC | o.class=ClassC @ o.i \}\also
\forall i:instances(ClassA) @  (\exists_1 x : ClassSpecA @ x.i=i)\\
\forall i:instances(ClassB) @  (\exists_1 x : ClassSpecB @ x.i=i)\\
\forall i:instances(ClassC) @  (\exists_1 x : ClassSpecC @ x.i=i)\\
\end{schema}

\noindent Now, the following schema defines how the system gets updated because of object creation :
\begin{schema}{NewA}
\Delta ObjectSystemABC\\
n?:ClassSpecA
\where
ObjectsA'=ObjectsA \cup \{n?\}
\end{schema}
\noindent Notice how we added to the schema ObjectSystemABC an attribute $ObjectsA:\power ClassSpecA$. This paragraph and the associated schemas should be taken as a parenthesis since our goal here is just to specify the global properties of an object system. We will hence continue using axiomatic definitions instead of schemas for the global object system, which makes most descriptions lighter and easier to read, as long as we do not plan to model how the system state can change.

\subsection{dereferencing attributes}
An essential operation in object systems is to obtain the information held by the data structure pointed at by an object reference. This operation, called "dereferencing" can me modeled in our case on a per attribute basis. We prefix the attribute name by the string "get", and promote the first attribute letter to upper case to name the accessor ("power" becomes "getPower"). Following is an example in our ABC toy problem :
\begin{axdef}
getA:A \fun \nat
\where
\forall i:A @ \\
\t1 getA(i)= (\mu v:\{s:ClassSpecA | s.i=i @ s.a\}\cup \{s:ClassSpecC | s.i=i @ s.a\}@v)
\end{axdef}

\noindent This definition uses Z's \emph{mu} construct $(\mu x : T | C @ E)$ that yields the value of $E$ on the unique $x$ from $T$ matching $C$. Again, it can be seen as a little verbose, as a result of Z's non object orientedness. However, it is easily specified, and such definitions can be generated automatically from shortcut descriptions.

\subsection{making a class abstract}

Now, based on the same example, if we expect the class $A$ to be abstract, i.e. we forbid an individual to be created as an $A$ we simply need to add a constraint stating that $instances(ClassX)$ is empty : $instances(ClassA)=\{o : ClassSpecA | class=ClassA @ o.i \}=\emptyset$. 

\subsection{unused objects}
The specification made so far accepts that elements of $ObjectReference$ are members of none of the subtypes. Depending upon the situation (e.g. whether a constraint programming tool using the specification must try giving a type to all the elements in $ObjectReference$ or not), we may force objects to belong to types. This is obtained by adding the axiom 
\begin{zed}
	\langle instances(ClassA), instances(ClassB), instances(ClassC) \rangle \partition ObjectReference
\end{zed} 

\subsection{specializing across several discriminators}
An important concept in object oriented specification is the possibility to specialize a class across two different discriminators, each corresponding to different viewpoints over a class. For instance, a traditional real life example is the class \emph{Vehicle}. It can be specialized in one discriminator, called "energy", related to the energy used to power the vehicle. We may imagine the subclasses \emph{Human}(powered), \emph{Wind}(powered), \emph{Gas}(powered) in that discriminator. Each subclass in this case brings its own data attributes  : number of humans, number of sails, tank capacity. The  \emph{Vehicle} class can also be specialized across another discriminator : the element it moves on. We can imagine here the classes : \emph{Water}, \emph{Ground}, \emph{Air}. Again, each of these classes may carry some data, in isolation from the others. In the declaration of types, it suffices to state the following (everything irrelevant has been removed):

\begin{axdef}
Vehicle, Human, Wind, Gas, Water, Ground, Air:ReferenceSet\\
\where
Vehicle = Human \cup Wind \cup Gas\\
Vehicle = Water \cup Ground \cup Air\also

instances(ClassHuman)=\{o : ClassSpecHuman | o.class=ClassHuman @ o.i \} \dots \also

\forall i:instances(ClassHuman) @  (\exists_1 x : ClassSpecHuman @ x.i=i) \dots
\end{axdef}

\noindent The rule in the UML is that whenever such a multiple discriminator specialization occurs, the main class (here $Vehicle$) and all its child classes (i.e. $Human, \ldots Air$) are abstract, and that any concrete class underneath must inherit at least one class from each discriminator. This is so because since vehicle is partitioned in two discriminators, any "Vehicle" must belong to some type among each discriminator. Obtaining this requires that each subclass inherits a class from each discriminator. The predicate stated in the axiomatic definition above ensure this: any object reference in a "sub"subclass of Vehicle must be a member of at least one set among $Human, Wind, Gas$, and of at least one set among $Water, Ground, Air$. Membership to those sets is acquired through inheritance.

\subsection{shortcut notation for class specifications}
Z being non object oriented in any way, the previous class and type declarations are verbose. For simplicity and readability, although not sacrificing rigor, we propose the following shortcut definition for classes and types, which makes use of the keywords \emph{class}, \emph{abstract}, \emph{discriminator}, \emph{inherit}. The syntax for this can be presented using simple examples, which must be understood as a shortcut for the corresponding specifications.   
\begin{schema}{class-A:abstract}
	-discriminators:default\\
	a:\nat
	\where
	a<10;
\end{schema}
\vspace{-0,7cm}
\begin{schema}{class-B:concrete}
	-inherit:A-default\\
	b:\nat_1
	
\end{schema}
\vspace{-0,7cm}
\begin{schema}{class-C:concrete}
	-inherit:A\\
	\where
	a\ge 5;
\end{schema}

\noindent A preprocessor can very easily parse such definitions, or take its input from an UML class design, so as to produce a listing identical to what was built step by step for the $ABC$ example in the previous pages. Hence, a byproduct of these declarations is the declaration in the Z specification of the schemas : $ObjectDef$, $ClassDefA$, $ClassSpecA$, $ClassDefB$, $\dots$, and of the sets $instances(ClassA)$, $A$, $instances(ClassB)$, $\dots$.

In the case of the $Vehicle$ class, since it has two discriminators, we would declare (all irrelevant information is hidden):

\begin{schema}{class-Vehicle:abstract}
	-discriminators:powermode,element\\
\end{schema}
\vspace{-0,7cm}
\begin{schema}{class-Human:abstract}
	-inherit:Vehicle-powermode\\
\end{schema}
\vspace{-0,7cm}
\begin{schema}{class-Ground:abstract}
	-inherit:Vehicle-element\\
\end{schema}
\vspace{-0,7cm}
\begin{schema}{class-Bicycle:concrete}
	-inherit:Human|Ground\\
\end{schema}

\section{Relations}
Z provides the richest possible toolkit to define relations and reason about them. This feature is inherent in relational languages, where all common mathematical concepts, like functions, bags, sequences derive from relations through composition and constraints. For instance, a \emph{function} is a relation bound by an axiom of unicity. Also, a \emph{sequence} is a function from a subset of natural numbers $\nat$ to a given set. 
\subsection{a simple example}
Having defined the structure and inheritance relations between classes, we must now describe their relations. Like before, we will study this through a concrete example, based on two classes $Person$ and $Company$. 
\begin{schema}{class-Person}
\end{schema}
\vspace{-0,7cm}
\begin{schema}{class-Company}
\end{schema}

This specification defines schemas: 
\begin{zed}
	ClassDefPerson \defs [~\dots | \dots~]\\
	ClassDefCompany \defs [~\dots | \dots~]\\
	ClassSpecPerson \defs [~\dots | \dots~]\\
	ClassSpecCompany \defs [~\dots | \dots~]\\
\end{zed}
as well as the appropriate constraints on $instances(ClassPerson), instances(ClassCompany)$ and also the type sets:
\begin{axdef}
	Person, Company: ReferenceSet\\
\end{axdef}
\subsection{relations and roles}

A relation is declared between types, no matter what the creation type of the objects is. In our example, we may think about these three relations :

\begin{axdef}
worksFor : Person \rel Company\\
owns : Person \rel Company\\
manages : Person \rel Company\\
\end{axdef}

\noindent In standard object oriented modeling\cite{UML}, relation names are complemented by role names, associated with each extremity of a class relation. Each role name names the target class role wrt. the particular relation. Role names must be specified by the object model. When not ambiguous (i.e. when only one relation binds two given classes), the target class name is implicitly accepted as a role name. Roles of binary relations can be axiomatically defined as follows :
\begin{axdef}
employees : Company \fun \power Person\\
employer : Person \fun \power Company
\where
\forall c : Company @ employees(c)= \{p : Person | (p \mapsto c) \in worksFor\}\\
\forall p : Person @ employer(p)= \{c : Company | (p \mapsto c) \in worksFor\}
\end{axdef}
Note that the Z syntax allows more compact definitions for the roles $employer$ and $employees$ : 
\begin{zed}
employees(c)=\dom (worksFor \rres \{c\})\\
employer(p)=\ran (\{p\}\dres worksFor)\\
\end{zed}
or also
\begin{zed}
employees(c)=worksFor \inv \limg \{c\} \rimg\\
employer(p)=worksFor \limg \{p\} \rimg
\end{zed}

\noindent where $worksFor \inv$ denotes the relational inverse of $worksFor$, $worksFor \dres \{c\}$ denotes the domain restriction of $worksFor$ wrt. $\{c\}$ (which is still a relation), $\dom R$ denotes the domain of $R$,  and $\_ \limg \_ \rimg$ is the relational image operator. 
We may ease the pain of declaring roles for all the relations in a model by generically defining the $lrole$ and $rrole$ Boolean functions as follows.
\begin{gendef}[C,D]
lrole:\power ((C \rel D) \cross (D \fun \power C))\\
rrole:\power ((C \rel D) \cross (C \fun \power D))
\where
\forall R : C \rel D; l : D \fun \power C @ (R,l) \in lrole \iff (\forall d : D @ l(d)=R\inv \limg \{d\} \rimg)\\
\forall R : C \rel D; r : C \fun \power D @ (R,r) \in rrole \iff (\forall c : C @ r(c)=R \limg \{c\} \rimg)
\end{gendef}

\noindent The previous axiomatic definition is \emph{generic}, parameterized with types (this is the first time we use this). Now, the declaration of the roles associated  to the relation $worksFor$ can be simplified :
\begin{axdef}
employees : Company \fun \power Person\\
employer : Person \fun \power Company\\
\where
(worksFor,employees)\in lrole\\
(worksFor,employer)\in rrole
\end{axdef}

\noindent It is also possible to generically (pre)define two roles for any arbitrary binary relation as follows : 
\begin{gendef}[p,c]
leftRole: (p \rel c) \cross c \fun \power p\\
rightRole: (p \rel c) \cross p \fun \power c
\where
\forall R : p \rel c; vc:c @ leftRole(R,vc)=R\inv \limg \{vc\} \rimg\\
\forall R : p \rel c; vp:p @ rightRole(R,vp)=R \limg \{vp\} \rimg\\
\end{gendef}

\noindent These definitions illustrate the amazing power of Z for defining builtin syntax extensions, as well as the richness of the relational operator toolkit of Z, later useful for specifying object constraints. It must be noted that from our viewpoint, Z offers a clear advantage over other object oriented\cite{Mailharro:98}, or terminological languages\cite{baader91terminological, Stumptner97} for object oriented constraints wrt. a potential broad acceptance, since relation definition is not role centered, but relations, functions and roles can freely coexist.

\subsection{composition, aggregate relations}

Object modeling leads to a clear separation between two broad categories of relations. General relations are unconstrained, meaning that every tuple can be accepted, regardless of the number of times an object appears on either side. For instance, in modeling a network of PCs and printers, any PC can view any number of printers (even though it may not see all of them), and any printer can be accessed by any number of PCs. No limitation stems from the nature of the relation itself. 

Other relations are more constrained. For instance, no PC can share its mainboard. This is an example of a \emph{composition} relationship. To distinguish between both just involves changing the type of the relation to make it a function of a special kind. If a relation stated between the composite type and the component type (in this sequence) is a composition one, it means that its relational inverse is an injective partial function (each component occurs in at most one composite). If no component can be left aside, the relational inverse is injective. Z provides various notations for constrained functions : injections start with an arrow ($\inj$, $\pinj$), surjections end with a double arrow  ($\surj$, $\psurj$), partial functions have a bar in the middle ($\pinj$, $\psurj$), bijections are both injective and surjective ($\bij$), whereas unconstrained functions are denoted with a simple arrow ($\fun$) and standard relations have two opposed arrows ($\rel$).


\begin{axdef}
uses : PC \rel Printer\\
hasMainBoard : PC \rel MainBoard\\
\where
hasMainBoard \inv \in MainBoard \pinj PC
\end{axdef}

\noindent If components cannot be optional, the injection becomes non partial 

\begin{axdef}
hasDVDWriter : PC \rel DVDWriter\\
\where
hasDVDWriter \inv \in DVDWriter \inj PC
\end{axdef}

\noindent In the most constrained case, of a strict one to one relation between types, the relation becomes a bijection, which can be formulated as follows :
\begin{axdef}
hasMainBoard : PC \bij MainBoard\\
\end{axdef}

\noindent More generally, any constraint can be stated upon a relation using general quantified formulas and all of Z's constructs. The distinction made in the UML between aggregate and standard relations is conceptual, and does not relate to constraints in our sense here. Aggregate relations models relations where a dynamic, not structural dependency exists among between objects. For instance, translating a paragraph in a text amounts to translating all its characters.

\subsection{multiplicities}
Relation multiplicities can be naturally stated as well. Object models often constrain for a given relation the number of related target objects for each source object. For instance, a PC has at most four memory units plugged (the $\#$ operator denotes a set cardinality). 

\begin{axdef}
hasMemory : PC \rel Memory\\
\where
\forall pc:PC @ \# (~hasMemory \limg \{pc\} \rimg~) \leq 4
\end{axdef}

\subsection{ordered relations}
Object models sometimes require that the tuple ordering is significant to a relation. For instance, should we model the relation between polygons and points, it is clear that we need a list, not a set, of points to describe the Polygon. The concept available in Z to model this is the sequence.
\begin{axdef}
builds : Polygon \fun (\seq Point)\\
\end{axdef}
To restrict the multiplicity in this case (for instance to describe all the pentagons) requires a little different work than before 

\begin{axdef}
builds : Polygon \fun (\seq Point)\\
\where
\forall p:Polygon @\# builds =5
\end{axdef}

\noindent To ensure that an object does not occur twice or more in a sequence, we need to make the sequences injective:

\begin{axdef}
builds : Polygon \fun (\iseq Point)\\
\end{axdef}

\noindent To decide that a Point in our example does not occur in the definition of two or mode different Polygons, we state :
\begin{axdef}
builds : Polygon \fun (\iseq Point)\\
\where
\forall p_1,p_2:Polygon @ items~(builds(p_1)) \cap items~(builds(p_2)) = \emptyset
\end{axdef}

\subsection{reified associations}
An important feature in object oriented modeling is the possibility to attach extra information to associations in the model. This added information is carried by an \emph{association class}, which can be a standard class (i.e. with a name) or be anonymous. We can however assume the existence of a name since the association class for a given relation $R$ can be named automatically (e.g. as $R\_DATA$)). 

We thus expect the association class used in coordination with a given relation to be properly defined as a class according to the former framework. If we return to the $worksFor$ example, we see that an obvious related information can be the salary (the salary can be different if a person works for two different companies, hence it cannot be an information carried by the Person itself).

\begin{schema}{class-EnrolmentInfo:concrete}
	salary:\nat
	\where
	a>MIN\_SALARY;
\end{schema}

\noindent This definition yields as usual two schemas : $ClassDefEnrolmentInfo$, $ClassSpecEnrolmentInfo$, and two sets : $instance(ClassEnrolmentInfo)$ and $EnrolmentInfo$. 
The latter is the type associated with objects built as members of $EnrolmentInfo$ itself or subclasses. Binding the enrolment information to the $worksFor$ relation is the fact of a function from the $worksFor$ tuples to the type EnrolmentInfo.

\begin{axdef}
mapEnrolmentInfo:worksFor \fun EnrolmentInfo
\end{axdef}

\noindent If the attached information is optional, the function is partial :
\begin{axdef}
mapOptionalEnrolmentInfo:worksFor \pfun EnrolmentInfo
\end{axdef}

\section{Constraints}

\subsection{structural constraints : example}
Constrained object oriented problems abound with constraints spanning across the object structure, traversing relations to gather information.  The Z notation again offers many possible ways to state such constraints. Lets study an example, classical in the configuration community. The model describes all valid PC's, composed from standard components, in a simplified form. 

We declare the following classes : $PC$, $PowerSupply$, $MainBoard$, $Monitor$, $Processor$. Except for $PowerSupply$ and $PC$ all the classes inherit an abstraction called $Device$, with an attribute called $powerUsed$. $PowerSupply$ has an attribute called $power$. We also have the relations $PC\_PowerSupply$, $PC\_MainBoard$, $PC\_Monitor$ and $MainBoard\_Processor$. The shortcut definitions for these classes are :

\begin{schema}{class-Device:abstract}
	powerUsed:\nat
\end{schema}
\vspace{-0,7cm}
\begin{schema}{class-PC:concrete}
\end{schema}
\vspace{-0,7cm}
\begin{schema}{class-PowerSupply:concrete}
	power:\nat
\end{schema}
\vspace{-0,7cm}
\begin{schema}{class-MainBoard:concrete}
	-inherit:Device\\
\end{schema}
\vspace{-0,7cm}
\begin{schema}{class-Processor:concrete}
	-inherit:Device\\
\end{schema}
\vspace{-0,7cm}
\begin{schema}{class-Monitor:concrete}
	-inherit:Device\\
\end{schema}
We also declare the composition relations (assuming default role names)
$PC\_PowerSupply$, $PC\_MainBoard$, $PC\_Monitor$ and $MainBoard\_Processor$\footnote{we intentionally continue to read the relations from composite to components to emphasize the fact that composition is a constraint}


\begin{axdef}
PC\_PowerSupply : PC \rel PowerSupply\\
PC\_Monitor : PC \rel Monitor\\
PC\_MainBoard : PC \rel MainBoard\\
MainBoard\_Processor : MainBoard \rel Processor\\
\where
PC\_PowerSupply \inv \in PowerSupply \inj PC\\
PC\_Monitor \inv \in Monitor \inj PC\\
PC\_MainBoard \inv \in MainBoard \inj PC\\
MainBoard\_Processor \inv \in Processor \inj MainBoard
\end{axdef}

\noindent Now, we wish to state the constraint that the total power delivered by a PowerSupply must exceed the total power demand by all the devices in the PC.
This is a classical example of structural constraint. To achieve this, we must define several utilities.

\subsection{structural constraints utilities}
We need several intermediate definitions useful to declare constraints. For instance, some integer arithmetic functions generalize to the case of bags of natural numbers, or numerals. Some of the properties of object systems require to gather some information over the structure (like the \emph{price}, or the \emph{power} used by electrical units for instance). Such information is best represented in bags, which allow repeated occurrences of the same value. Given a bag, we may ask for its min, max, or sum for instance. We detail these three function, which may serve as a template for possible others. 

In the rest of the sub section, we also provide a function that can be used to generate a sequence from a set. Since converting from bags to sets is immediate, and converting from sequences to bags is predefined by the function $items$, this allows to convert any structure type into any other.
\subsubsection*{computing the min and max over a bag of naturals}
\begin{axdef}
bagmin:\bag \nat \fun \nat\\
bagmax:\bag \nat \fun \nat
\where
\forall b:\bag \nat @ bagmin(b)=min~ (\dom b)\\
\forall b:\bag \nat @ bagmax(b)=max~ (\dom b)
\end{axdef}

\subsubsection*{summing up the elements in a bag of naturals}
\begin{axdef}
bagsum:\bag \nat \fun \nat
\where
bagsum(\emptyset)=0\\
\forall b:\bag \nat | (\dom b)\neq \emptyset@\\
\t1 (\LET x==bagmin(b)@ bagsum(b)=b(x)*x + bagsum(b \setminus \{(x,b(x))\}))
\end{axdef}
To understand this definition of $bagsum$, it suffices to recall that a bag is a partial function from a set to strictly positive integers (the number of times an element is counted).

\subsubsection*{conversion functions}

We define a conversion function from totally ordered finite sets to sequences. This function $asSeq$ converts a finite set to a sequence, which may be further converted to a bag using the $items$ operator on sequences. This provides full possibilities of converting from a container type to another. We need a function to select a member from a set. This is possible deterministically for totally ordered sets, as are $\nat$ or the set of rational numbers. We present the specification of the conversion function $asSeq$ in the case of natural numbers. This gives a template for the definition of similar conversion functions applying to totally ordered sets of non integral elements. 

\begin{axdef}
asSeq:\finset \nat \fun \seq \nat
\where 
asSeq(\emptyset)=\langle \rangle\\
\forall S : \finset_1 \nat @ (\LET x == (max~S) @ asSeq(S)=asSeq(S \setminus \{x\}) \cup \{\#S \mapsto x\})
\end{axdef}

\subsubsection*{building a bag of attribute values}
Together with any attribute, we know that we can specify an accessor function which, given an object reference $i$, will return the attribute value held by the object structure mapped to $i$. We have established the convention of naming $getXyz$ the accessor function for attribute $xyz$. We generalize this concept to sets of object references. We want, given a set of object references, to build the bag of corresponding attribute values. In our example, we expect the following accessor functions to be implicitly defined from the class declaration as follows :





\begin{axdef}
getPower:PowerSupply \fun \nat
\where
\forall i:Device @ getPower(i)= (\mu v: \{s:ClassSpecPowerSupply | s.i=i @ s.power\}@v)
\end{axdef}

\begin{axdef}
getPowerUsed:Device \fun \nat
\where
\forall i:Device @ getPowerUsed(i)= \\
\t1 (\mu v: \{s:ClassSpecMonitor | s.i=i @ s.powerUsed\} \cup  \\
\t1 \{s:ClassSpecProcessor | s.i=i @ s.powerUsed\} \cup  \\
\t1  \{s:ClassSpecMainBoard | s.i=i @ s.powerUsed\ \} @ v )
\end{axdef}

\noindent We further make the assumption that the set ObjectReference to be totally ordered\footnote{This is not unrealistic, since object references generally will be interpreted as integers (machine pointers \emph{are} integers).}, which allows to define a function called $pickFirst$ yielding the first element of any finite set of ObjectReferences :
\begin{axdef}
pickFirst: \finset ObjectReference \fun ObjectReference
\end{axdef}

\noindent From these accessors and the function $first$, we may form their generalized counterpart as (we use $bag$ as a prefix to form the function names) :
\begin{axdef}
bagPower:\finset PowerSupply \fun \bag \nat
\where
bagPower(\emptyset)=\lbag \rbag\\
\forall d:\finset_1 PowerSupply  @ (\LET x == pickFirst(d) @ \\
\t1 bagPower(d) =  (bagPower(d\setminus \{x\}) \uplus (\{getPower(x) \mapsto 1\})))
\end{axdef}
In the same spirit, we could define $bagPowerUsed$ by simply replacing $PowerSupply$ by $Device$, and $getPower$ by $getPowerUsed$ in the previous statement.

\begin{axdef}
bagPowerUsed:\finset Device \fun \bag \nat\\
\dots
\end{axdef}

\noindent As in the case of association roles, it is possible to elaborate a generic definition for these :

\begin{gendef}[X]
bagOf: (ObjectReference \fun X) \fun (\finset ObjectReference \fun \bag X )
\where
\forall f:ObjectReference \fun X @ bagOf(f)(\emptyset)=\lbag \rbag\\
\forall f:ObjectReference \fun X @ \\
\t1 \forall d:\finset_1 (\dom f)  @ (\LET x == pickFirst(d) @ \\
\t1 bagOf(f)(d) =  (bagOf(f)(d\setminus \{x\}) \uplus (\{f(x) \mapsto 1\})))
\end{gendef}
The function $bagOf$ hence maps every function from $ObjectReference$ to $X$ to a function from sets of $ObjectReference$ to bags of $X$.

\subsection{inter relation constraints}
The basic constraint existing between relations is the subset constraint. A simple example is given by the two relations $worksFor$ and $manages$, between the types $Person$ and $Company$. The manager obviously works for the company, which is expressed as $manages \subset worksfor$. Hence the proper declaration for these relations becomes :

\begin{axdef}
worksFor : Person \rel Company\\
manages : Person \rel Company\\
\where
manages \subset worksFor
\end{axdef}

\subsection{structural constraints}

We wish to state the constraint that the total power delivered by the power supply suffices to feed all of the PC's devices. This can be stated as follows :

\begin{zed}
\forall p:PC @\\
(\LET R == PC\_Monitor \cup PC\_MainBoard \cup MainBoard\_Processor @ \\
\t1 bagsum(bagOf(getPower)(PC\_PowerSupply \limg \{p \} \rimg )) \geq \\
\t1 bagsum(bagOf(getPowerUsed)(R \plus \limg \{p \} \rimg \cap Device))  )\\
\end{zed}
where $R \plus$ denotes the transitive closure of the relation $R$, obtained as the union of three relations, and $R \plus \limg \{p \} \rimg$ denotes the relational image of $p$, the PC composite, by $R\plus$, hence the component objects of $p$, at any structural level.

\subsection{notational shortcuts for relations and roles}


Most often, specifications require to make the structure traversal more explicit. To illustrate the possibilities offered by Z in that respect, we assume that all previous relations have roles named using a standard prefix "the" followed by the distant class name (we use no $s$ at the end, even when there can be several), as e.g. :
\begin{axdef}
theMonitor: \finset PC \fun \finset Monitor \\
\end{axdef}

\begin{gendef} [X]
\_ \rightarrow \_ : \finset ObjectReference \cross (ObjectReference \fun X) \fun \bag X\\
\_ \rightharpoonup \_ : \finset ObjectReference \cross (ObjectReference \fun X) \fun X\\
\_ \cdot \_ : \finset ObjectReference \cross (\finset ObjectReference \fun \finset X) \fun \finset X\\
\_ \leadsto \_ :  ObjectReference \cross (\finset ObjectReference \fun \finset X) \fun \finset X\\
\where
\forall s:\finset ObjectReference; r: ObjectReference \fun  X @ s \rightarrow r = bagOf(r)(s)\\
\forall s:\finset ObjectReference; r: ObjectReference \fun  X @ s \rightharpoonup r = (\mu t: bagOf(r)(s)@first~t)\\
\forall s:\finset ObjectReference; r:\finset ObjectReference \fun \finset X @ s \cdot r = r(s)\\
\forall o:ObjectReference; r:\finset ObjectReference \fun \finset X @ o \leadsto r = r(\{o\})\\
\end{gendef}

\begin{zed}
\forall p:PC @\\
\t1 p \leadsto thePowerSupply \rightharpoonup getPower \geq \\
\t1 p\leadsto theMonitor \rightharpoonup getPowerUsed + \\
\t1 p\leadsto theMainBoard \rightharpoonup getPowerUsed + \\
\t1 bagsum(p\leadsto theMainBoard \cdot theProcessor \rightarrow getPowerUsed) \\
\end{zed}

\section{An AI Example}

\begin{figure}[htb]
   \centering
      \includegraphics[angle=270,width=\textwidth]{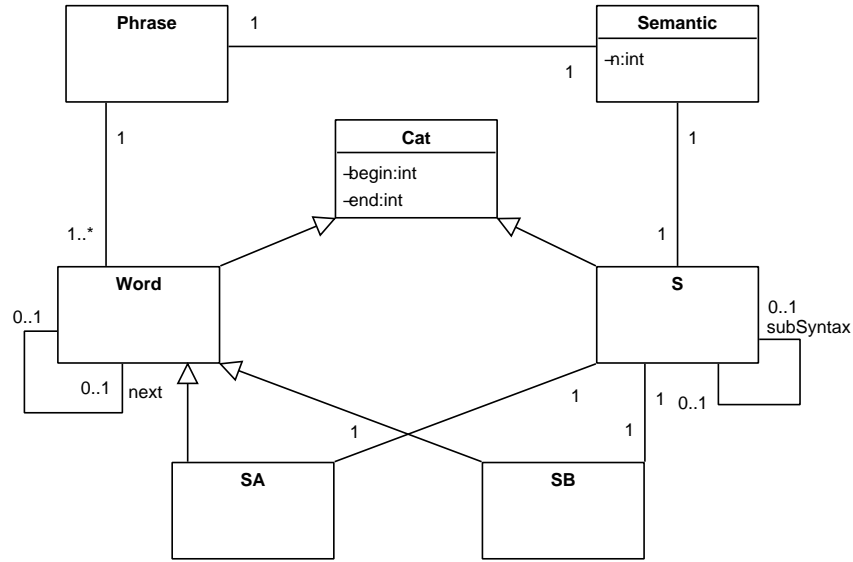}
    \caption{An object model for the $a^nb^n$ parsing problem}
    \label{fig:anbn}
\end{figure}

We wish to illustrate the use of the specification utilities and frameworks presented so far with a simple yet very general artificial intelligence problem. \cite{estratat2003} defines the problem of analyzing both the syntax and semantic of a context free language using a constraint object system. the language chosen is the archetypal language $\mathcal{L} = a^n~b^n$ consisting in sequences of a's followed by the same number of b's. $aaabbb \in \mathcal{L}$, but $abbb \notin \mathcal{L}$. \cite{estratat2003} proposes the object model and constraints described below to represent valid sentences of $\mathcal{L}$ together with their semantic. The object model is illustrated in figure \ref{fig:anbn}. The program used to solve the problem is Ilog JConfigurator, an object oriented configurator. Given an input made of a sequence of words, some of them not being classified as a's or b's, the system can generate all the valid word sequences compatible with that input together with the correct syntax structure. The system works equally well when chunks of syntactical structure, or elements of the semantic, are fed in.

In other words, the system produces the following results (where inputs and outputs are sequences $\langle string, syntax\ tree, semantic \rangle$) (we use the dot character "." do denote an unknown word (a, or b), and the character "?" to denote an unknown :
\begin{zed}
\langle aaabbb, ?, ?\rangle \mapsto \langle aaabbb,S(SA,S(SA,S(SA,null,SB),SB),SB),3\rangle\\
\langle abbb, ? , ? \rangle  \mapsto false\\
\langle .~a~.~b, ?, ? \rangle  \mapsto \langle aabb,S(SA,S(SA,null,SB),SB),2\rangle\\
\langle ?, ?, 2 \rangle  \mapsto \langle aabb,S(SA,S(SA,null,SB),SB),2\rangle\\
\end{zed}

We propose here a rigorous, type checked specification of the object model and its constraints, that illustrate the power of the method. We start with the definition of several classes. $Word$\footnote{Following the terminology of natural language theories, we use "phrase" to denote a valid sentence for the grammar, called a "word" or a "string" in formal language theory} is an abstraction for $SA$ and $SB$ (representing a and b), $Cat$ is an abstraction for both $Word$ and $S$. $S$ is the only syntactic construct, made of an $SA$, an optional enclosed $S$, and an $SB$ in that order. The $Phrase$ consists of a list of $Word$, a syntax $S$, and a $Semantic$. The semantic chosen is as simple as the example : it describes the count of a's and b's in the sentence.

\begin{schema}{class-Phrase:concrete}
\end{schema}
\vspace{-0,7cm}
\begin{schema}{class-Cat:abstract}
begin:\nat\\
end:\nat
\end{schema}
\vspace{-0,7cm}
\begin{schema}{class-Word:abstract}
- inherit: Cat
\end{schema}
\vspace{-0,7cm}
\begin{schema}{class-S:concrete}
-inherit: Cat
\end{schema}
\vspace{-0,7cm}
\begin{schema}{class-SA:concrete}
-inherit: Word
\end{schema}
\vspace{-0,7cm}
\begin{schema}{class-SB:concrete}
-inherit: Word
\end{schema}
\vspace{-0,7cm}
\begin{schema}{class-Semantic:concrete}
n:\nat
\end{schema}


\noindent Several relations exist in this problem. They can be described very naturally by their most used roles. Whenever the opposite role is needed, the inverse of the relation can be computed. Each phrase maps to a unique first word. Each word maps to a unique phrase. Each word has an optional next word. Each phrase bijectively maps to a semantic. It also maps to a unique syntax S. Each SA (and SB) bijectively maps to an S. Each S has an optional enclosed S (we use a partial injection here). Each S is in a one to one with its semantic. We also know that the first word is a member of the phrase words. All these elements can be formulated as :
\begin{axdef}
firstWord : Phrase \fun Word\\
phrase : Word \fun Phrase\\
next: Word \pinj Word\\
phraseSemantic: Phrase \bij Semantic\\
phraseSyntax : Phrase \fun S\\
SASyntax:SA \bij S\\
SBSyntax:SB \bij S\\
subSyntax:S \pinj S\\
semantic: S \bij Semantic
\where
firstWord \subset phrase \inv
\end{axdef}

\noindent In some constraints below, we expect the following functions to be implicitly defined : 
\begin{axdef}
theSA : \finset S \fun \finset SA\\
theSB : \finset S \fun \finset SB\\
theSubS :  \finset S \fun \finset S\\
thePhraseSyntax :  \finset Phrase \fun \finset S\\
\end{axdef}

\noindent The following accessor functions are also implicitly defined :

\begin{axdef}
getBegin:Cat \fun \nat\\
getEnd:Cat \fun \nat\\
getN:Semantic \fun \nat\\
\end{axdef}

Using these definitions, we may formulate the following axioms, necessarily verified by the object system. Of course, some or all of these axioms must be implemented as constraints in a working system.

\noindent The length of words is one
\begin{zed}
\forall w:Word@ getBegin(w) +1 = getEnd(w)\\
\end{zed}

\noindent The start position of the first word in a phrase is 0
\begin{zed}
\forall p:Phrase@ getBegin(firstWord(p))=0\\
\end{zed}

\noindent The first word in a phrase is the SA of its syntax (S)
\begin{zed}
\forall p:Phrase@ firstWord(p)=SASyntax \inv(phraseSyntax(p))\\
\end{zed}

\noindent Consecutive words have corresponding end/begin
\begin{zed}
\forall w:\dom next @ getEnd(w) = getBegin(next(w))\\
\end{zed}

\noindent All a's are located left of all b's
\begin{zed}
\forall a:SA;b:SB@ getBegin(a) < getBegin(b)\\
\end{zed}

\noindent The beginning of an S is the beginning of its SA (and respectively with SB's and "ends").
\begin{zed}
\forall s:S @ getBegin(s) = s\leadsto theSA  \rightharpoonup getBegin\\
\forall s:S @ getEnd(s) = s\leadsto theSB \rightharpoonup getEnd\\
\end{zed}

\noindent The enclosed S is between the SA and the SB.
\begin{zed}
\forall s:\dom subSyntax @ getBegin(s) < s\leadsto theSubS  \rightharpoonup getBegin\\
\forall s:\dom subSyntax @ getEnd(s) > s\leadsto theSubS \rightharpoonup getEnd\\
\end{zed}

\noindent The end position of the SA plus the length of the enclosed S equals the start position of the SB
\begin{zed}
\forall s:\dom subSyntax@ \\
\t1 s\leadsto theSA \rightharpoonup getEnd + s\leadsto theSubS  \rightharpoonup getEnd - s\leadsto theSubS  \rightharpoonup getBegin =  \\
\t1 s\leadsto theSB \rightharpoonup getBegin\\
\forall s:S | s \notin \dom subSyntax@ s\leadsto theSA \rightharpoonup getEnd  = s\leadsto theSB \rightharpoonup getBegin\\
\end{zed}
The semantic of a phrase is the semantic of its syntax
\begin{zed}
\forall p:Phrase@ phraseSemantic(p) = semantic(phraseSyntax(p))\\
\end{zed}
The "value" of the semantic of an "S" is the integer division of the its length by two
\begin{zed}
\forall s:S@ getN(semantic(s)) = (getEnd(s)-getBegin(s)) \div 2\\
\forall s:S@ getN(semantic(s)) \mod 2 = 0\\
\end{zed}

\noindent Not all these axioms are independent of course. However, they formally describe all the valid object configurations that are instances, or solutions, of this object model. Provided the class definitions are properly expanded, or this expansion is simulated, all the constraints can be fully type checked by a Z type checker. Furthermore, these axioms can be input to a theorem prover, with the possibility of generating automatic or user assisted proofs for conjectures about the properties of the problem, or proofs that some constraints are mutually incompatible.

The same specification may also be converted automatically to a valid input for any practical configurator or object constraint program.

\subsection{Formal proofs}
All object models so specified naturally allow formal proofs to be made about the axiom set. Essential in that respect are redundancy proofs. In constraint systems, any axiom that can provably be inferred from the rest of the axioms can be safely ignored by an implementation, which hence remains correct. Also, redundant axioms can be added when their implementation as a constraint has a better propagation efficiency than the axioms it can be derived from. In this case, redundancy ensures that the resulting system remains complete. 

We illustrate the possibility to establish formal proofs for our example. $\forall s:S@ getN(semantic(s)) \mod 2 = 0$ can be proved by induction on the height of the syntactical structure (or the value of the semantic "n"). A sequent proof for height $1$ (ie. for an "S" having no enclosed "S", or in other words for the "S" corresponding to the central "AB") is :
\begin{infrule}
\forall s:S | s \notin \dom subSyntax@ s\leadsto theSA \rightharpoonup getEnd  = s\leadsto theSB \rightharpoonup getBegin\\
\forall w:Word@ getBegin(w) +1 = getEnd(w)
\derive
\forall s:S| s \notin \dom subSyntax@ getEnd(s)-getBegin(s) = 2
\end{infrule}

\begin{infrule}
\forall s:S| s \notin \dom subSyntax@ getEnd(s)-getBegin(s) = 2\\
\forall s:S@ getN(semantic(s)) = (getEnd(s)-getBegin(s)) \div 2\\
\derive
\forall s:S| s \notin \dom subSyntax@ getN(semantic(s)) \mod 2 = 0\\
\end{infrule}

\noindent Now, we can formally prove that if the induction hypothesis is true for height $n$, it holds for height $n+1$, hence for all $n$. We can first establish as a lemma that the length of an $S$ equals $2$ plus that of its subSyntax :

\begin{infrule}\forall s:\dom subSyntax@ \\
\t1 s\leadsto theSA \rightharpoonup getEnd + s\leadsto theSubS  \rightharpoonup getEnd - s\leadsto theSubS  \rightharpoonup getBegin =  \\
\t1 s\leadsto theSB \rightharpoonup getBegin\\
\forall w:Word@ getBegin(w) +1 = getEnd(w)
\derive
\forall s:\dom subSyntax@ getEnd(s)-getBegin(s) = \\
\t1 2+s\leadsto theSubS  \rightharpoonup getEnd - s\leadsto theSubS  \rightharpoonup getBegin
\end{infrule}
which makes the proof of the induction step obvious.

\section{Conclusion}

We have presented the entire specification of an object oriented constraint system, which can be used to document and exchange constrained object models. We used Z as an underlying formal system which offers many advantages:
\begin{itemize}
\item Z has very simple and clean first order semantics.
\item as a relational language, Z offers the richest possible ways of reasoning about relations, which is an essential aspect of constrained object systems.
\item Z is freely extensible by introducing new operators, always backed by rigorous axiomatic definitions. This allows to attain the flexibility and readability of existing object oriented approaches.
\item the Z language comes with a freely available type checker \fuzz, that allows to control both the syntax and the type conformance of specifications (this article is fully type checked using \fuzz).
\end{itemize}
Our goal was to capture as much of object oriented modeling as possible, using as a basis the widely accepted standard UML, while avoiding to produce a new avatar of an object oriented language. We also rejected the idea of using an existing one, since all existing formal object languages have their pros and their cons, which might have interfered with the general objective of producing a tool for communicating and discussing constrained object systems. Even though some of our choices may still be discussed, this can be made formally. Furthermore, the Z language being formally extensible at will, all the proposed generic operators can be viewed as a template, rather than a rule.

\bibliography{2003rr-lsis-03-006}

\begin{thebibliography}{10}

\bibitem{ooze91}
Antonio Alencar and Joseph Goguen.
\newblock Ooze: An object-oriented {Z} environment.
\newblock In {\em Pierre America, editor, European Conference on Object
  Oriented Programming, Springer, Lecture Notes in Computer Science, Volume
  512}, pages 180--199, 1991.

\bibitem{AFM02}
J\'er\^ome Amilhastre, H\'el\`ene Fargier, and Pierre Marquis.
\newblock Consistency restoration and explanations in dynamic csps--application
  to configuration.
\newblock {\em Artificial Intelligence}, 135(1-2):199--234, 2002.

\bibitem{baader91terminological}
F.~Baader and B.~Hollunder.
\newblock A terminological knowledge representation system with complete
  inference algorithms.
\newblock In {\em Proceedings of the First International Workshop on Processing
  Declarative Knowledge}, volume 572, pages 67--85, Kaiserslautern (Germany),
  1991. Springer--Verlag.

\bibitem{SS-ZUM91}
Rosalind Barden, Susan Stepney, and David Cooper.
\newblock The use of {Z}.
\newblock In J.~E. Nicholls, editor, {\em Proceedings of the 6th Z User
  Meeting, York, UK, 1991}, Workshops in Computing, pages 99--124. Springer,
  1992.

\bibitem{barkerconnor89}
Virginia Barker, Dennis O'Connor, Judith Bachant, and Elliot Soloway.
\newblock Expert systems for configuration at digital: Xcon and beyond.
\newblock {\em Communications of the ACM}, 32:298--318, 1989.

\bibitem{borba94operational}
P.~Borba and J.~Goguen.
\newblock An operational semantics for {FOOPS}.
\newblock In R.~Wieringa and R.~Feenstra, editors, {\em Working Papers of the
  International Workshop on Information Systems - Correctness and
  Reusabilility, {IS}-{CORE}'94. Technical Report {IR}-357}, Amsterdam, 1994.

\bibitem{klone85}
R.~J. Brachman and J.~G. Schmolze.
\newblock An overview of the kl-one knowledge representation system.
\newblock {\em Cognitive Science}, 9(2):171 -- 216, 1985.

\bibitem{caseau94constraint}
Yves Caseau.
\newblock Constraint satisfaction with an object-oriented knowledge
  representation language.
\newblock {\em Applied Intelligence}, 4(2):157--184, 1994.

\bibitem{estratat2003}
Mathieu Estratat.
\newblock Application de la configuration \`a l'analyse syntaxico s\'emantique
  de descriptions.
\newblock Master's thesis, Facult\'e des Sciences et Techniques de Saint
  J\'er\^ome, Marseille, France, 2003.

\bibitem{Felfernig02}
Alexander Felfernig, Gerhard Friedrich, Dietmar Jannach, and Markus Zanker.
\newblock Semantic configuration web services in the cawicoms project.
\newblock In {\em Proceedings of the Configuration Workshop, 15th European
  Conference on Artificial Intelligence}, pages 82--88, Lyon, France, 2002.
\newblock http://www.cawicoms.org/.

\bibitem{FSB99}
Markus P.~J. Fromherz, Vijay~A. Saraswat, and Daniel~G. Bobrow.
\newblock Model-based computing: Developing flexible machine control software.
\newblock {\em Artificial Intelligence}, 114(1-2):157--202, October 1999.

\bibitem{UML}
Object~Management Group.
\newblock {\em UML v. 1.5 specification}.
\newblock OMG, 2003.

\bibitem{jacobs99coalgebras}
B.~Jacobs.
\newblock Coalgebras in specification and verification for objectoriented
  languages, 1999.

\bibitem{JL87}
Joxan Jaffar and Jean~Louis Lassez.
\newblock Constraint logic programming.
\newblock In {\em in ACM Symposium on Principles of Programming Languages},
  pages 111--119, 1987.

\bibitem{JohnGeske99}
Ulrich John and Ulrich Geske.
\newblock Reconfiguration of technical products using conbacon.
\newblock In {\em Proceedings of AAAI'99-Workshop on Configuration}, pages
  48--53, Orlando, Florida, July 1999.

\bibitem{Mailharro:98}
Daniel Mailharro.
\newblock A classification and constraint-based framework for configuration.
\newblock {\em AI in Engineering, Design and Manufacturing, (12)}, pages
  383--397, 1998.

\bibitem{McDermott82}
John~P. McDermott.
\newblock R1: A rule-based configurer of computer systems.
\newblock {\em Artificial Intelligence}, 19:39--88, 1982.

\bibitem{Mittal:90AA}
Sanjay Mittal and Brian Falkenhainer.
\newblock Dynamic constraint satisfaction problems.
\newblock In {\em Proceedings of AAAI-90}, pages 25--32, Boston, MA, 1990.

\bibitem{Nareyek:99}
Alexander Nareyek.
\newblock Structural constraint satisfaction.
\newblock In {\em Papers from the 1999 AAAI Workshop on Configuration,
  Technical Report, WS-99-0}, pages 76--82. AAAI Press, Menlo Park, California,
  1999.

\bibitem{Meyer99}
Harald~Meyer nauf'm Hofe.
\newblock Construct: Combining concept languages with a model of configuration
  processes.
\newblock In {\em Papers from the 1999 AAAI Workshop on Configuration,
  Technical Report, WS-99-0}, pages 17--22, 1999.

\bibitem{Plain2002}
Kevin~R. Plain.
\newblock Optimal configuration of logically partitionned computer products.
\newblock In {\em Proceedings of the Configuration Workshop, 15th European
  Conference on Artificial Intelligence}, pages 33--34, Lyon, France, 2002.

\bibitem{sabinfreudercomposite96}
Daniel Sabin and Eugene~C. Freuder.
\newblock Composite constraint satisfaction.
\newblock In {\em Artificial Intelligence and Manufacturing Research Planning
  Workshop}, pages 153--161, 1996.

\bibitem{objectz}
Graeme Smith.
\newblock {\em The Object-Z Specification Language}.
\newblock Kluwer Academic Publishers, in Advances in Formal Methods, 2000.

\bibitem{soininen99fixpoint}
Timo Soininen, Esther Gelle, and Ilkka Niemela.
\newblock A fixpoint definition of dynamic constraint satisfaction.
\newblock In {\em Proceedings of CP'99}, pages 419--433, 1999.

\bibitem{soininen01representing}
Timo Soininen, Ilkka Niemela, Juha Tiihonen, and Reijo Sulonen.
\newblock Representing configuration knowledge with weight constraint rules.
\newblock In {\em Proceedings of the {AAAI} Spring Symp. on Answer Set
  Programming: Towards Efficient and Scalable Knowledge}, pages 195--201, March
  2001.

\bibitem{zspivey}
J.~M. Spivey.
\newblock {\em The Z Notation: a reference manual}.
\newblock Prentice Hall originally, now J.M. Spivey, 2001.

\bibitem{Stumptner97}
Markus Stumptner.
\newblock An overview of knowledge-based configuration.
\newblock {\em AI Communications}, 10(2), June 1997.

\bibitem{Stumptner98AIEDAM}
Markus Stumptner, Gerhard Friedrich, and Alois Haselböck.
\newblock Generative constraint-based configuration of large technical systems.
\newblock {\em Artificial Intelligence in Engineering, Design, Analysis and
  Manufacturing (AI EDAM)}, 12(4), Special Issue on Configuration, December
  1998.

\bibitem{Ylinen:02}
Katariina Ylinen, Tomi Männistö, and Timo Soininen.
\newblock Configuring software products with traditional methods - case linux
  familiar.
\newblock In {\em Proceedings of the Configuration Workshop, 15th European
  Conference on Artificial Intelligence}, pages 5--10, Lyon, France, 2002.

\end{thebibliography}
\bibliographystyle{plain}

\label{lastpage}
\end{document}